\documentclass[12pt]{article}

\usepackage{amsmath}
\usepackage{amssymb}
\usepackage{float}
\usepackage[T1]{fontenc}
\usepackage{graphicx}
\usepackage{caption}
\usepackage[utf8]{inputenc}
\usepackage{mathtools}
\usepackage[normalem]{ulem}
\usepackage{wasysym}
\usepackage[hidelinks]{hyperref}

\title{Generating Photo-realistic Images from LiDAR Point Clouds with Generative Adversarial Networks}

\begin{document}
\maketitle

\begin{scriptsize}
\begin{center}
\textbf{Nuriel Shalom Mor, Ph.D. \\ }\textit{Darca and Bnei Akiva Schools, Israel \\}
\end{center}

\begin{center}
\textbf{Jonathan Rabinowitz, Ilya Fridburg, Roi Beker, and Liran Zuriel \\ }\textit{AI Program Graduates, Hadash Darca School, Israel \\ }
\end{center}

\begin{center}
\textbf{Yoav Asnapi, and Eliyahu Aicho \\ }\textit{AI Program Students, Harel Bnei Akiva School, Israel \\}
\end{center}

\begin{center}
\textbf {Omer David Keilaf, Nir Goren, Raz Biderman, Amir Day, Oded Handel, and Yossi Avivi \\ }\textit{Innoviz Technologies LTD}
\end{center}

\end{scriptsize}

\begin{small}

\textbf{Abstract.} We examined the feasibility of generative adversarial networks (GANs) to generate photo-realistic images from LiDAR point clouds. For this purpose, we created a dataset of point cloud–image pairs and trained the GAN to predict photo-realistic images from LiDAR point clouds containing reflectance and distance information. Our models learned how to predict realistically looking images from just point cloud data, even images with black cars. Black cars are difficult to detect directly from point clouds because of their low level of reflectivity. This approach might be used in the future to perform visual object recognition on photo-realistic images generated from LiDAR point clouds. In addition to the conventional LiDAR system, a second system that generates photo-realistic images from LiDAR point clouds would run simultaneously for visual object recognition in real-time. In this way, we might preserve the supremacy of LiDAR and benefit from using photo-realistic images for visual object recognition without the usage of any camera. In addition, this approach could be used to colorize point clouds without the usage of any camera images. 
\end{small}

\section{Introduction}

Laser scanned (LiDAR) point clouds create a 3D map of the surroundings and provide a 360-degree view that helps the car drive in any type of condition. The LiDAR system determines at the highest level of accuracy, the distance, and range of an object, providing much-needed data to self-driving cars. LiDAR surpasses cameras in many aspects. However, it seems that visual recognition is one domain that might be done more easily with cameras, especially of black objects (Cui et al., 2021; Fernandes et al., 2021). Black objects are difficult to detect from point clouds because of their low level of reflectivity.

\begin{flushleft}
The purpose of this research was to examine whether conditional generative adversarial networks (cGANs) could generate accurate photo-realistic images from LiDAR point clouds data. We wanted to investigate if this approach could be used to apply object recognition and semantic segmentation on photo-realistic images generated from LiDAR point clouds. In this approach, we might preserve the supremacy of LiDAR and benefit from using photo-realistic images for visual object recognition and semantic segmentation. In addition, this approach potentially could be used to colorize point clouds without the usage of any camera images.   
\end{flushleft}

\begin{flushleft}
LiDAR point clouds are sparse and are difficult for the task of photo-realistic rendering (Cui et al., 2021; Fernandes et al., 2021).  Our approach to tackle this problem was to create a large dataset of point cloud–image pairs and train the cGAN to predict photo-realistic views from LiDAR point clouds which contain reflectance and distance information. The idea was that the model would learn how to predict realistically looking images from just point cloud data. 
\end{flushleft}

\subsection{Generative Adversarial Networks (GANs)}

\begin{flushleft}
GANs are the state-of-the-art framework for training generative models. Backpropagation is used to train, no inference is required during learning avoiding the difficulty of approximating intractable probabilistic computations (Creswell et al., 2018; Mirza $\&$ Osindero, 2014). 
\end{flushleft}

\begin{flushleft}
GANs consist of two adversarial neural networks:. 1. a generative network that captures the data distribution, and 2. a discriminative model that estimates the probability, a sample is from the training data rather than the generator (Isola,  Zhou, $\&$ Efros, 2017).  The generator and the discriminator can be deep neural networks and more specifically convolutional neural networks (CNNs; Isola et al., 2017). The domain of computer vision using CNNs is one of the areas that progressed dramatically during the deep learning era with many different types of applications (Mor $\&$ Dardeck, 2021; Mor $\&$ Dardeck, 2018). 
\end{flushleft}

\begin{flushleft}
The generator and the discriminator are trained simultaneously to adjust both parameters following a two-player min-max game. The generator produces images and the task of the discriminator is to distinguish between real images and images that were generated by the generator.  The aim of the discriminator is to assign the correct label to the real example and the example generated by the generator (real vs. generated). The aim of the generator is to  ``outsmart" the discriminator, so it would label generated images as real (Creswell et al., 2018; Mirza $\&$ Osindero, 2014). 
\end{flushleft}

\begin{flushleft}
\textit{\textbf{pix2pix network. }pix2pix }is a cGANs able to perform high-resolution image to image translation. This network learns the mapping from an input image to an output image and a loss function to train this mapping (Isola et al., 2017). Conditional GAN is a type of GAN in which we can direct the data generation process based on extra information such as class labels (Atienza, 2019; Peters, $\&$ Brenner, 2020).  
\end{flushleft}

\begin{flushleft}
The purpose of this research was to examine whether \textit{pix2pix }could generate accurate photo-realistic images from LiDAR point clouds given sufficient training data. 
\end{flushleft}

\section{Method}

\subsection{Data Acquisition}

\begin{flushleft}
The points were acquired with a state-of-the-art LiDAR -  InnovizOne:  \\ angular resolution of 0.1x0.1; configurable frame rate of 5-20 FPS; detection range of 0.1 - 250m; maximum field of view of 115x25 (Innoviz, 2021).  
\end{flushleft}

\begin{flushleft}
InnovizOne has four individually-controlled regions of interest (one on each laser) for dynamic focus in a limited FOV. This allows for enhanced visibility and range with no impact on bandwidth, resolution, or frame rate. InnovizOne returns multiple reflections per pixel and records and stores multiple points in a 3D environment. This is important when laser pulses hit rain droplets, snowflakes, or more than one object on its path. InnovizOne has no gaps in its scanning pattern due to contiguous pixels, which is critical to building a safe autonomous vehicle perception system. Without this, a system could miss collision-relevant small objects lying on the road’s surface or humans if they are within point cloud data gaps, causing devastating results. Provides uniform resolution over the entire vertical FOV. This enables the system to obtain more data than other sensors that focus only on the center (horizon) and lose data moving towards the edges. The InnovizOne vertical FOV is also built with panning capabilities to support mounting tolerances, as well as varying driving conditions such as vehicle loading. InnovizOne is resilient to ambient light sources, such as direct sunlight and blinding lights from oncoming cars, as well as poor weather conditions such as rain (Innoviz, 2021).
\end{flushleft}

\subsection{Preparing the data and training using pix2pix}

\begin{flushleft}
The dataset contained about 30,000 images and about 30,000 x 200,000 3D points.  \textit{pix2pix} needs the data in terms of a point cloud image and a corresponding real image.  We conducted two separate experiments. In experiment 1, the point cloud image contained one channel of the reflectance of the laser ray. In experiment 2, the point cloud image contained two channels. The first channel was the distance between a 3D point and the LiDAR and the second channel was the reflectance of the laser ray. 
\end{flushleft}

\begin{flushleft}
The task was therefore that the point cloud image would capture the same view and information as the real image. Inspired by the work of Peters, and Brenner (2020), we extracted the front view orientation of the 3D point clouds and stored in one channel the reflectivity and in the second channel the distance (instead of the pixel value). With this approach, we created point clouds images suitable as an input for training a GAN such as \textit{pix2pix}. The point cloud images and the real images of the same view enables us to examine the purpose of this research: whether cGAN such as pix2pix could generate accurate photo-realistic images from LiDAR point clouds data.  
\end{flushleft}

\begin{flushleft}
By the end of the process, we had about 30,000  point cloud images and corresponding 30,000 real images. In experiment 1 the point cloud images stored reflectivity and in experiment 2 the point cloud images stored reflectivity in one channel and distance in a second channel. 
\end{flushleft}

\begin{flushleft}
Our training procedure was inspired by \textit{pix2pix} with the required technical modifications and adjustments to fit our unique dataset (Isola et al., 2017). \\ 
\end{flushleft}

\section{Results And Experiments}

\subsection{Experiment 1}

\begin{flushleft}
The input point cloud image contained one channel of reflectivity. We trained the GANs for 50 epochs with a batch size of one. After 50 epochs the generator started to overfit the real images. Figures 1 and 2 show an example of predicted images from the test set containing black cars. The test set is a completely new recording that the GANs have never seen before the test.  This was predicted using only reflectivity information from the point cloud. We selected to show frames with black cars because black cars are usually difficult to detect from LiDAR. We can see that the generator learned to generate black cars, probably from contextual information, because of the fact that the colors and the exact shapes of objects in predicted images are not identical as in the real images. 
\end{flushleft}

\begin{figure}[H]
\raggedright
\includegraphics[width=11.38cm,height=7.06cm]{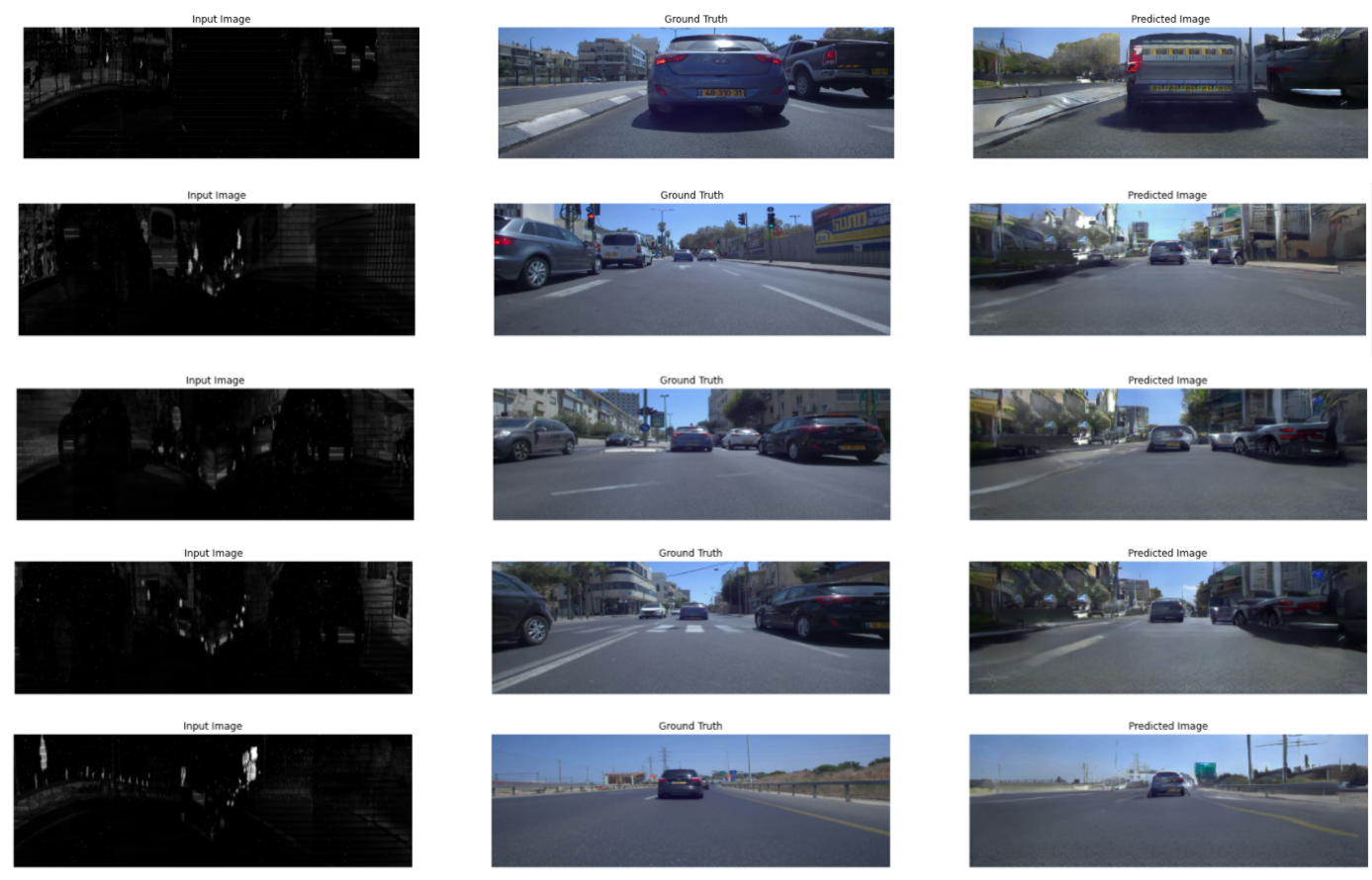}
\caption{Examples with black cars from test set of experiment 1}
\label{fig:examples_black_cars_test_set}
\end{figure}

\begin{figure}[H]
\raggedright
\includegraphics[width=11.38cm,height=7.06cm]{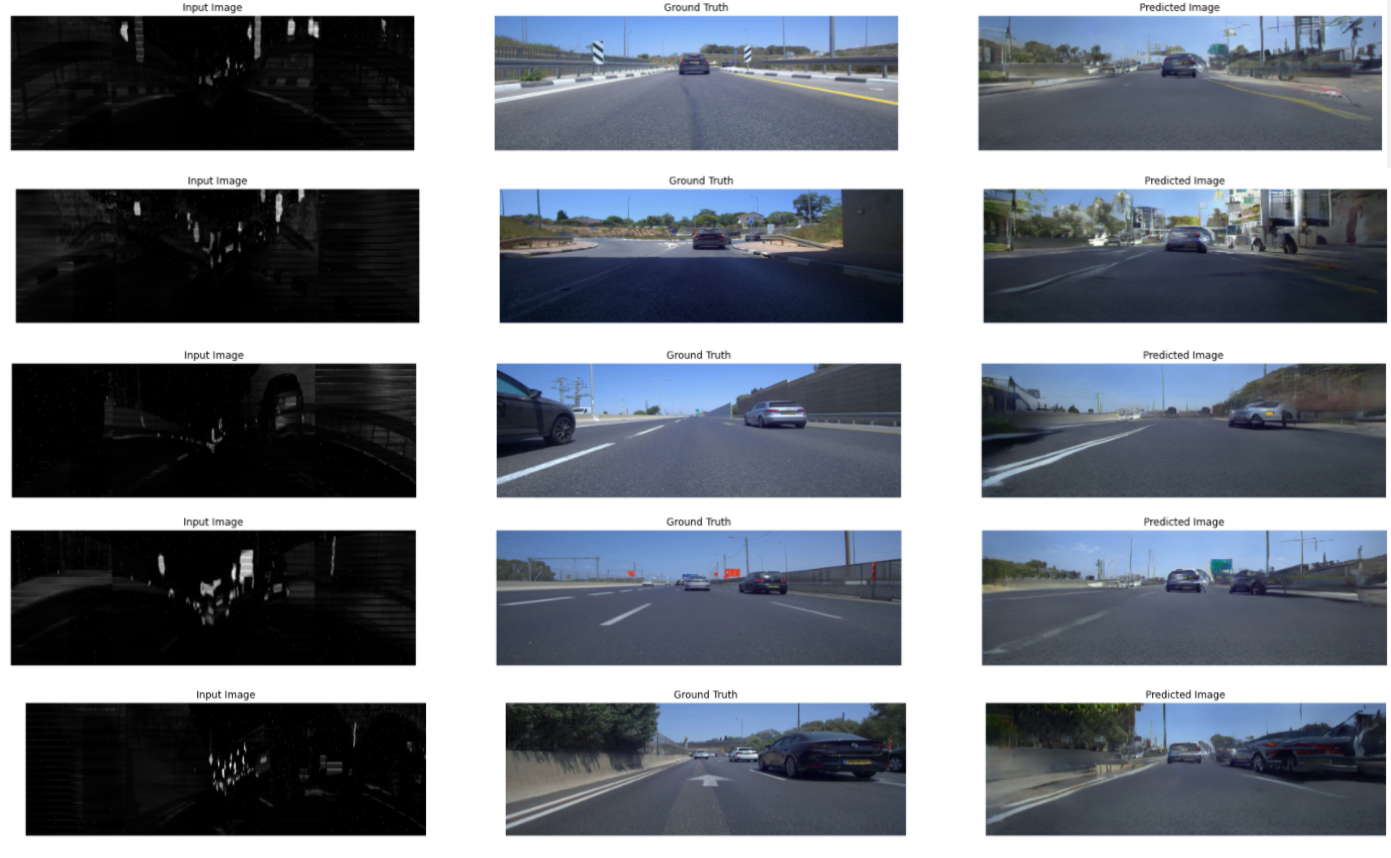}
\caption{Examples with black cars from test set of experiment 1}
\label{fig:examples_black_cars_test_set}
\end{figure}

\subsection{Experiment 2}

The input point cloud image is composed of two channels one of reflectivity and the second of distance. We trained the GANs for 40 epochs with a batch size of one. After 40 epochs the generator started to overfit the real images.  Figures 3 and 4 show an example of predicted images from the test set containing black cars. The test set is a completely new recording that the GANs have never seen before the test.  This was predicted using reflectivity and distance information from the point cloud. Also here we selected to show frames with black cars because it is challenging to detect them directly from LiDAR point clouds. Similar to experiment 1, we can see that the generator learned to generate black cars probably from contextual information, because of the fact that the colors and the exact shapes of objects in predicted images are not identical as in the real images. 

\begin{figure}[H]
\raggedright
\includegraphics[width=11.38cm,height=9.22cm]{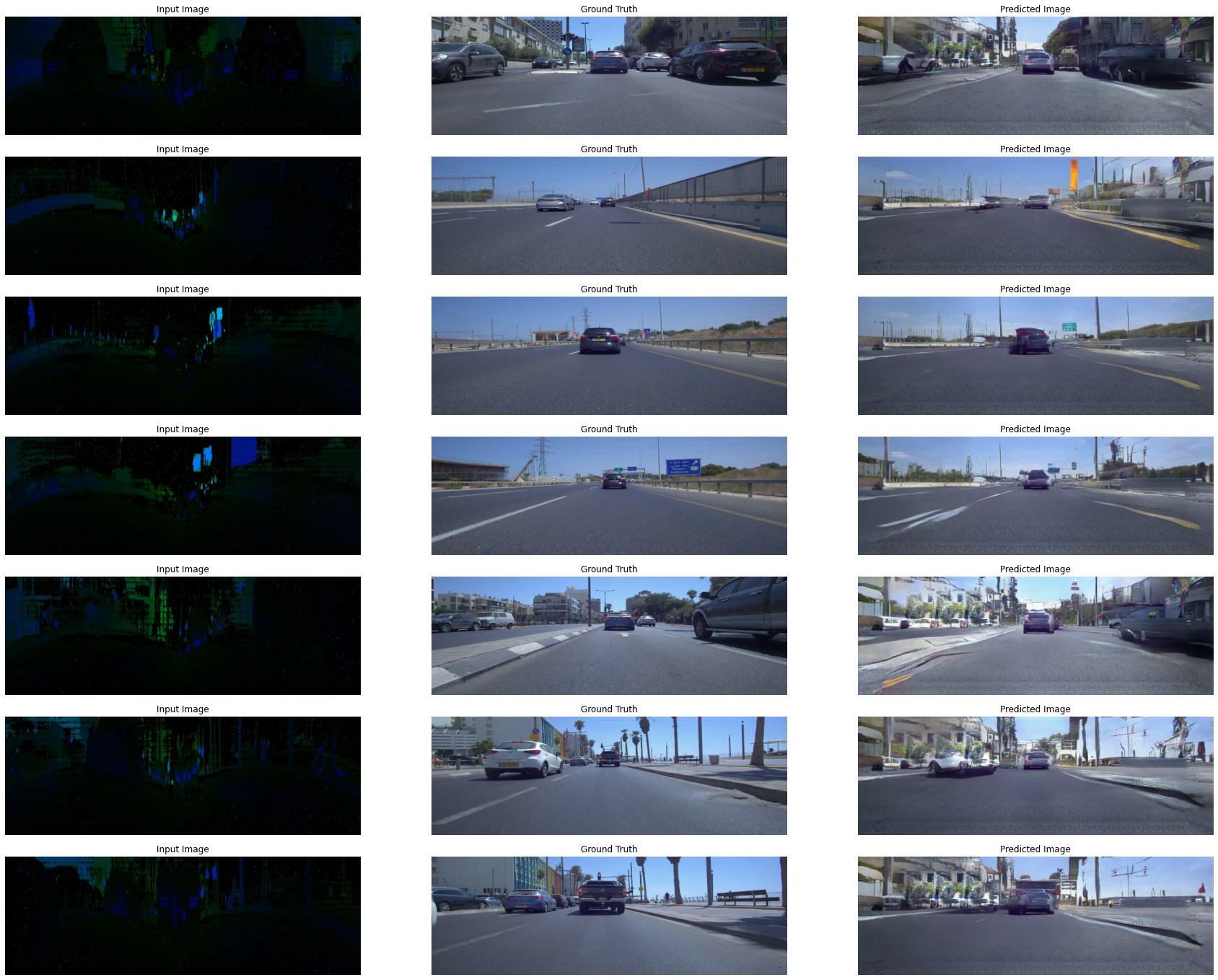}
\caption{Examples with black cars from test set of experiment 2}
\label{fig:examples_black_cars_test_set}
\end{figure}

\begin{figure}[H]
\raggedright
\includegraphics[width=11.38cm,height=7.9cm]{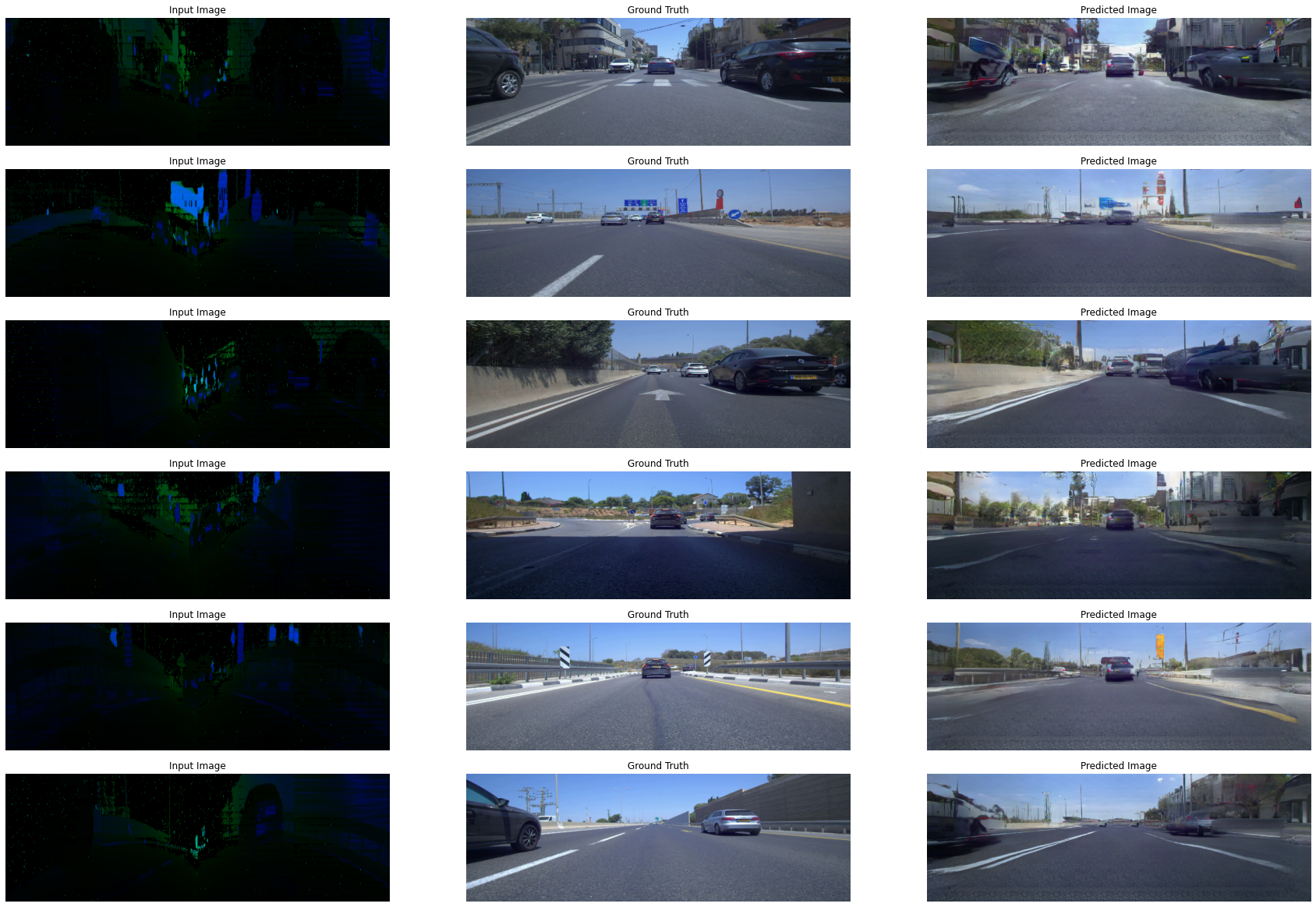}
\caption{Examples with black cars from test set of experiment 2}
\label{fig:examples_black_cars_test_set}
\end{figure}

A sample video generated from test set which is a completely new recording that the GANs have never seen before the test:  \\ \href{https://drive.google.com/file/d/1k_VNPYM8j2fVbzVbhHnCI21N9rZQ9uIW/view?usp=sharing}{\uline{video\_test\_exp\_2.mp4}}

\subsection{Evaluation}

\begin{flushleft}
The task of evaluating GANs’ performance is very difficult. There is no objective function used when training GAN generator models, meaning models must be evaluated using the quality of the generated synthetic images (Alqahtani et al., 2019; DeVries et al., 2019). Manual inspection of generated images is a good starting point when getting started. Quantitative measures, such as the inception score and the Frechet inception distance can evaluate the quality of the generated images (Alqahtani et al., 2019; DeVries et al., 2019).
\end{flushleft}

\begin{flushleft}
Our purpose was to examine the similarity of each specific predicted image compared to its specific ground truth, in terms of objects present in the images, and not the general visual quality of predicted images. Therefore, we developed a measure for our specific purpose.  For each pair, we calculated the number of cars that could be detected in the predicted image divided by the number of cars present in the ground truth (the real image). We summed this term for all examples and divided it by the number of examples. The closer the score is to 1, the more accurate the predicted images in terms of cars that are present.
\end{flushleft}

\begin{equation}
\label{eq:nolabel}
\frac{1}{m}\sum \frac{n_p}{n_g}
\end{equation}

\begin{flushleft}
We selected 100 pairs from the test set of experiment 1 and of experiment 2. The score in both experiments was between 0.7 and 0.8. Considering the fact that the general quality of the predicted images is lower than the real images (it is more difficult in general to detect objects in lower quality images), this score indicates that the vast majority of cars that present in the ground truth present in the predicted images. 
\end{flushleft}

\section{Conclusion}

We have demonstrated that it is possible to generate photo-realistic images from LiDAR point cloud data. We described an efficient methodology to represent point clouds as an image for using them as an input to deep neural networks as \textit{pix2pix.} 

\begin{flushleft}
In addition, we demonstrated that black cars which are difficult to detect directly from LiDAR point clouds can be generated and detected more easily in predicted images (predicted from LiDAR point clouds). The fact that in predicted images, color information and exact shapes are not identical to ground truth, suggests that that prediction of black cars is mostly derived from contextual information and not from the LiDAR reflectivity of the points themselves. We suggest that, in addition to the conventional LiDAR system, a second system that generates photo-realistic images from LiDAR point clouds would run simultaneously for visual object recognition in real-time. 
\end{flushleft}

\begin{flushleft}
In future work, to improve models' performance and quality of the predicted images, a larger diverse training set is required.
\end{flushleft}

\section{Authors’ Note}
Dedicated to the memorial of my father, Moshe Mor, peace be upon him.
\begin{flushleft}
Please send correspondence to Nuriel S. Mor, Ph.D. Software engineering program with an emphasis on AI, Darca and Bnei Akiva schools, Israel. Email: \href{mailto:nuriel.mor@gmail.com}{\uline{nuriel.mor@gmail.com}}
\end{flushleft}
The idea for such a system, as described in this article, was first proposed and developed by Dr. Mor in July 2021

\begin{center}
\_\_\_\_\_\_\_\_\_\_
\end{center}

We wish to thank:
\begin{flushleft}
Innoviz’s CEO and co-founder, \textbf{Omer David Keilaf} for allowing high school students from Holon, Bat Yam, Israel, an opportunity to experiment with data acquired by state-of-the-art LiDAR - InnovizOne. 
\end{flushleft}
\begin{flushleft}
Innoviz's software engineer, \textbf{Yossi Avivi}, for his support.
\end{flushleft}

\section{References}

\begin{flushleft}
Alqahtani, H., Kavakli-Thorne, M., Kumar, G., $\&$ SBSSTC, F. (2019, December). An analysis of evaluation metrics of gans. In \textit{International Conference on Information Technology and Applications (ICITA)}.
\end{flushleft}

\begin{flushleft}
Atienza, R. (2019). A conditional generative adversarial network for rendering point clouds. In \textit{Proceedings of the IEEE/CVF Conference on Computer Vision and Pattern Recognition Workshops} (pp. 10-17).
\end{flushleft}

\begin{flushleft}
Creswell, A., White, T., Dumoulin, V., Arulkumaran, K., Sengupta, B., $\&$ Bharath, A. A. (2018). Generative adversarial networks: An overview. \textit{IEEE Signal Processing Magazine}, \textit{35}(1), 53-65.
\end{flushleft}

\begin{flushleft}
Cui, Y., Chen, R., Chu, W., Chen, L., Tian, D., Li, Y., $\&$ Cao, D. (2021). Deep learning for image and point cloud fusion in autonomous driving: A review. \textit{IEEE Transactions on Intelligent Transportation Systems}.
\end{flushleft}

\begin{flushleft}
DeVries, T., Romero, A., Pineda, L., Taylor, G. W., $\&$ Drozdzal, M. (2019). On the evaluation of conditional gans. \textit{arXiv preprint arXiv:1907.08175}.
\end{flushleft}

\begin{flushleft}
Fernandes, D., Silva, A., Névoa, R., Simões, C., Gonzalez, D., Guevara, M., ... $\&$ Melo-Pinto, P. (2021). Point-cloud based 3D object detection and classification methods for self-driving applications: A survey and taxonomy. \textit{Information Fusion}, \textit{68}, 161-191.
\end{flushleft}

\begin{flushleft}
Innoviz (2021). https://innoviz.tech/
\end{flushleft}

\begin{flushleft}
Isola, P., Zhu, J. Y., Zhou, T., $\&$ Efros, A. A. (2017). Image-to-image translation with conditional adversarial networks. In \textit{Proceedings of the IEEE conference on computer vision and pattern recognition} (pp. 1125-1134).
\end{flushleft}

\begin{flushleft}
Mirza, M., $\&$ Osindero, S. (2014). Conditional generative adversarial nets. \textit{arXiv preprint arXiv:1411.1784}
\end{flushleft}

\begin{flushleft}
Mor, N. S., $\&$ Dardeck, K. L. (2021). Applying a Convolutional Neural Network to Screen for Specific Learning Disorder. \textit{Learning Disabilities: A Contemporary Journal, 19}(2), 161-169.
\end{flushleft}

\begin{flushleft}
Mor, N. S., $\&$ Dardeck, K. L. (2020). Applying Deep Learning to Specific Learning Disorder Screening. \textit{arXiv preprint arXiv:2008.13525}.
\end{flushleft}

\begin{flushleft}
Mor, N. S., $\&$ Dardeck, K. L. (2018). Quantitative Forecasting of Risk for PTSD Using Ecological Factors: A Deep Learning Application. \textit{Journal of Social, Behavioral, and Health Sciences, 12}(1), 61-73. DOI:10.5590/JSBHS.2018.12.1.04. 
\end{flushleft}

\begin{flushleft}
Peters, T., $\&$ Brenner, C. (2020). Conditional adversarial networks for multimodal photo-realistic point cloud rendering. \textit{PFG–Journal of Photogrammetry, Remote Sensing and Geoinformation Science}, \textit{88}(3), 257-269.

\end{flushleft}

\end{document}